\documentclass[Afour,sageh,times]{sagej}

\usepackage{moreverb,url}
\usepackage{subfigure}
\usepackage[colorlinks,bookmarksopen,bookmarksnumbered,citecolor=red,urlcolor=red]{hyperref}

\newcommand\BibTeX{{\rmfamily B\kern-.05em \textsc{i\kern-.025em b}\kern-.08em
T\kern-.1667em\lower.7ex\hbox{E}\kern-.125emX}}

\begin{document}

\runninghead{Kokash et al.}

\title{Ontology Creation and Management Tools: the Case of Anatomical Connectivity}

\author{Natallia Kokash\affilnum{1}, Bernard de Bono\affilnum{2} and Tom Gillespie\affilnum{3}}

\affiliation{\affilnum{1}Institute of Informatics, University of Amsterdam, The Netherlands 
\affilnum{2}Auckland Bioengineering Institute, University of Auckland, New Zealand 
\affilnum{3}Department of Neuroscience, University of California, CA, USA}

\corrauth{Natallia Kokash, Institute of Informatics, University of Amsterdam, The Netherlands.}
\email{nkokash@uva.nl}

\begin{abstract}
We are developing infrastructure to support researchers in mapping data related to the peripheral nervous system and other physiological systems, with an emphasis on their relevance to the organs under investigation. The nervous system, a complex network of nerves and ganglia, plays a critical role in coordinating and transmitting signals throughout the body. To aid in this, we have created ApiNATOMY, a framework for the topological and semantic representation of multiscale physiological circuit maps. ApiNATOMY integrates a Knowledge Representation (KR) model and a suite of Knowledge Management (KM) tools. The KR model enables physiology experts to easily capture interactions between anatomical entities, while the KM tools help modelers convert high-level abstractions into detailed models of physiological processes, which can be integrated with external ontologies and knowledge graphs.
\end{abstract}


\maketitle

\section{Introduction}
\label{sect:introduction}

Ontologies are essential for developing standardized vocabularies and defining relationships that help describe and interpret data from diverse sources. They are crucial for achieving semantic interoperability in many domains, allowing different systems to exchange data with a consistent and shared meaning. Ontologies are extensively used in biological and biomedical research~\cite{hoehndorf2015role,bioKM}, due to their ability to: 
\begin{itemize}
    \item provide standard identifiers for classes and relationships representing complex  phenomena;
    \item include metadata to clarify the intended meaning of classes and relationships; 
    \item include machine-readable definitions that allow computational access to class properties and relationships;
    \item standardize vocabulary across multiple data sources.
\end{itemize}

Ontology-based data integration plays a vital role in neuroscience, where researchers synthesize knowledge across physiology, anatomy, molecular and developmental biology, cytology, and mathematical modeling to support accurate data representation, analysis, and simulation. 

A common challenge for many large neuroscience projects is the integration of data across a wide diversity of species, spatial resolutions, and temporal scales.
The Stimulating Peripheral Activity to Relieve Conditions (SPARC) program is a NIH-funded consortium to improve the understanding of how the Autonomic Nervous System (ANS) interacts with end organs and the Central Nervous System (CNS)~\cite{Sparc}. A key objective of SPARC is to leverage this knowledge to develop the next generation of neuromodulation devices as effective therapies for various diseases. To aid in neuromodulation planning, the program maps the pathways of ANS neuron populations, including those involved in visceral sensing, by identifying the anatomical structures in which they traverse or terminate. This map serves as an organizational framework for (i) SPARC's expanding repository of electro-physiological recordings and molecular assay data, and (ii) the computational simulation of neural stimulation effects on the CNS or ANS.

The SPARC map represents connectivity as a network of conduits called the SPARC Connectivity Knowledge Base of the Autonomic Nervous System (SCKAN). SCKAN encapsulates detailed knowledge about CNS-ANS-end organ circuitry, derived from SPARC data and scientific literature, and is structured to support computational reasoning. All connections are annotated with SPARC’s standard reference vocabularies, enabling the integration of datasets that adhere to the same annotation standards~\cite{surles2022extending}. The circuits depicted on the map detail ANS connectivity specific to various organs, such as those involved in bladder control, defensive breathing, modulation of peristalsis, or cardiac inotropy/chronotropy. These circuits are constructed using input from SPARC investigators, anatomical experts, and scientific literature. They include comprehensive representations of neuron populations responsible for ANS connections, detailing the locations of cell bodies, dendrites, axon segments, and synaptic endings.

The SPARC Knowledge Graph (KG) is an integrated graph database composed of three parts: 
\begin{itemize}
    \item the SPARC dataset metadata graph, 
    \item ApiNATOMY models of connectivity, 
    \item a combination of the Neuroscience Information Framework (NIF) ontology~\cite{gardner2008neuroscience} and community ontologies.
\end{itemize}
ApiNATOMY circuits~\cite{BSG+15} are augmented with general connectivity knowledge between CNS nuclei, ANS ganglia, nerves, and end organs, derived from the scientific literature. To facilitate the extraction and ongoing relevance of this knowledge from the literature, Natural Language Processing (NLP) pipelines is used to extract connectivity relationships between distinct anatomical structures within sentences from scientific texts~\cite{savova2019use}. SCKAN is utilized to create an interactive visual atlas of ANS circuits accessible through the SPARC portal. There are two representations of SCKAN that serve complementary use cases. The triple store is useful for executing basic competency queries over the dataset releases, but there are no existing APIs that are straightforward to consume. On the other hand, SciGraph provides a developer-friendly REST API that is easy to use in production systems~\cite{SciGraph}.

In this paper, we focus on the part of the framework that deals with the development of ApiNATOMY models of connectivity. On a technical level, the main question is how do we cope with the complexity of the accumulated data and validate their quality~\cite{roadmap2016}. A central element of our approach is the use of a visual toolkit that supports data integration in a multiscale connectivity model of cells, tissues, and organs. The ApiNATOMY toolset consists of KR and KM tools~\cite{ApiNATOMY-tool,ApiNATOMY-demo} that enable topological and semantic modeling of process routes and associated anatomical compartments in multiscale physiology. Domain experts provide model specifications in a semi-structured format based on templates for common topological structures such as, e.g., neuronal chains. ApiNATOMY tools then expand the specifications into a graph that can be visually manipulated, serialized, and integrated with other knowledge bases such as the NIF-Ontology and the SPARC KG~\cite{osanlouy2021sparc}.

ApiNATOMY models are annotated using the same vocabularies as datasets, allowing connectivity knowledge to be leveraged to enhance navigation, classification, and data search. For example, if a researcher is designing a procedure on the Middle Cervical Ganglion (MCG), ApiNATOMY connectivity provides the means to discover existing data along the route of neurons passing through the MCG that could help plan the experiment.

We present our work on the ApiNATOMY infrastructure and describe a workflow process adopted by field experts adding novel data to shared knowledge graphs. This includes an interactive graphical web application, services to support identifier resolution, and tools to query linked metadata, discover datasets and track provenance. We use a generalizable approach to make application-specific JSON~\cite{JSON} data structures FAIR (Findable, Accessible, Interoperable, Reusable)~\cite{wilkinson2016fair} by using JSON-LD~\cite{JSON-LD}, a lightweight Linked Data format, to convert them to Resource Description Framework (RDF)~\cite{RDF}. This approach has greatly reduced the complexity of the system and made it easier to maintain.


\section{ApiNATOMY: Purpose and Approach}\label{sect:apinatomy} 

Biomedical KM is a multidisciplinary discipline characterized by complex infrastructure, processes, and knowledge processing pipelines to ensure large-scale acquisition, representation, and quality assessment of relevant data. KM pipelines take care of placing curated and annotated data into public data repositories. As part of such effort, we deal with inter-organ connectivity knowledge acquisition and visualization to characterize neural pathways in the context of the anatomy of a species and render them as an aid to users navigating, discovering, retrieving, and understanding complex and varied datasets. 

\begin{figure}
  \centering
  \includegraphics[width=0.5\textwidth]{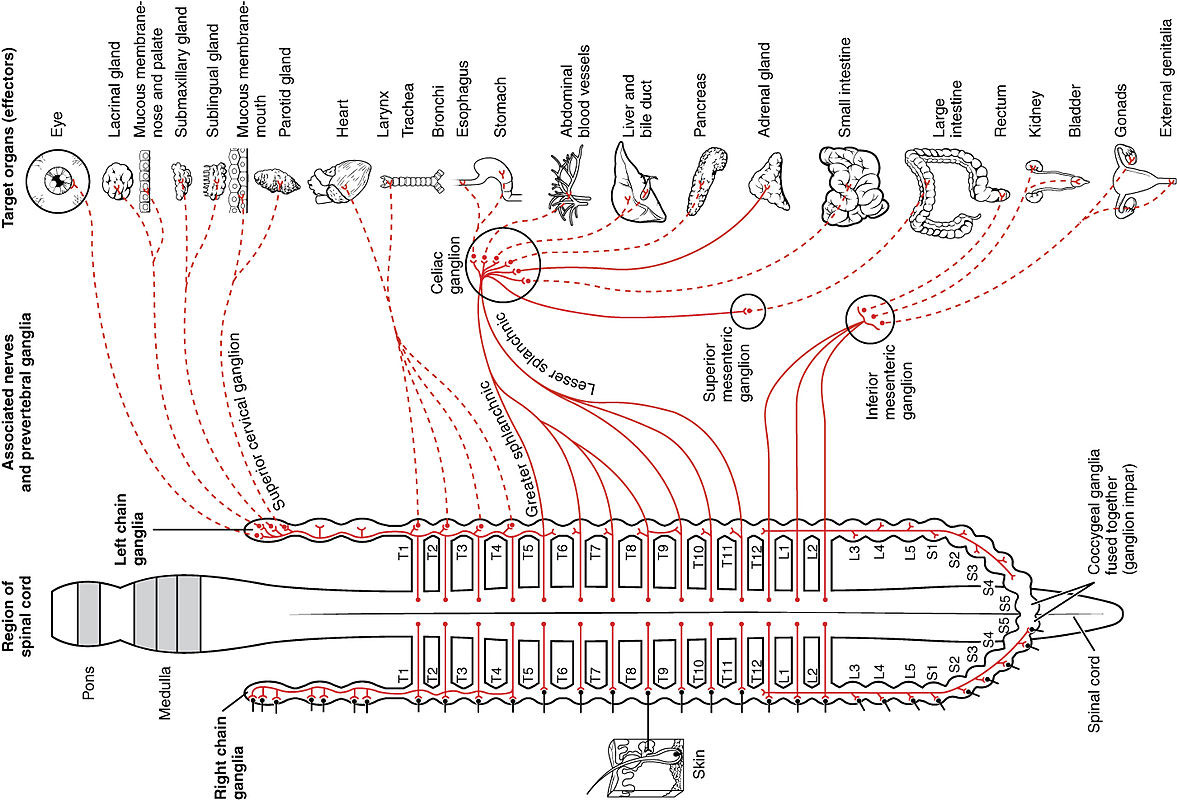}
\caption{Textbook illustration of the sympathetic nervous system with sympathetic cord and target organs}
  \label{fig:textbook}
\end{figure}

\begin{figure*}
  \centering
  \subfigure[Scale 1: Sympathetic nervous system]{
    \includegraphics[width=0.97\textwidth]{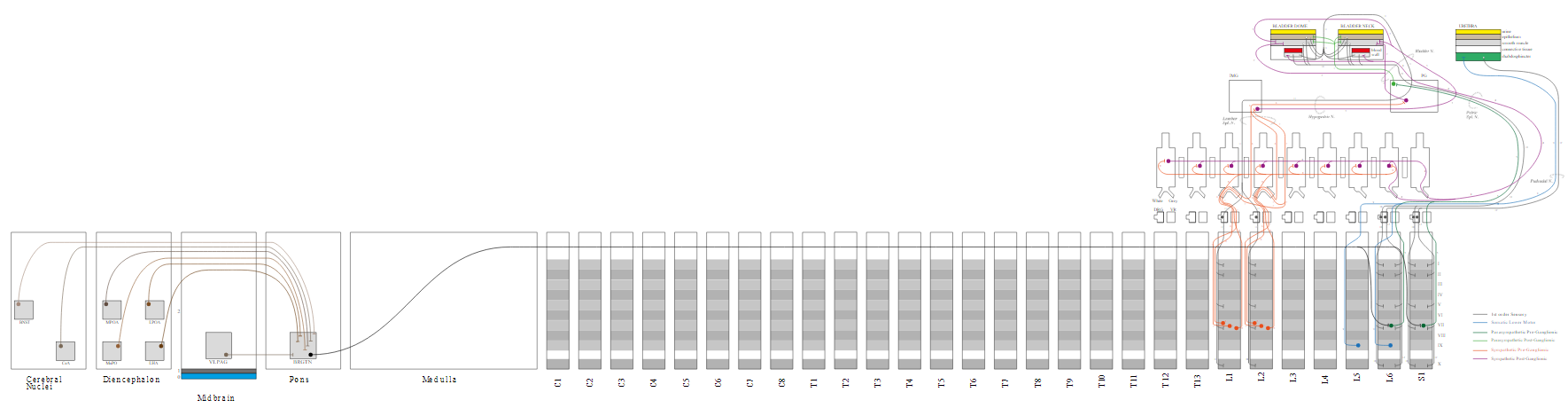}
    \label{fig:keastFull}
}
\subfigure[Scale 2: Bladder modeling]{
    \includegraphics[width=0.47\textwidth]{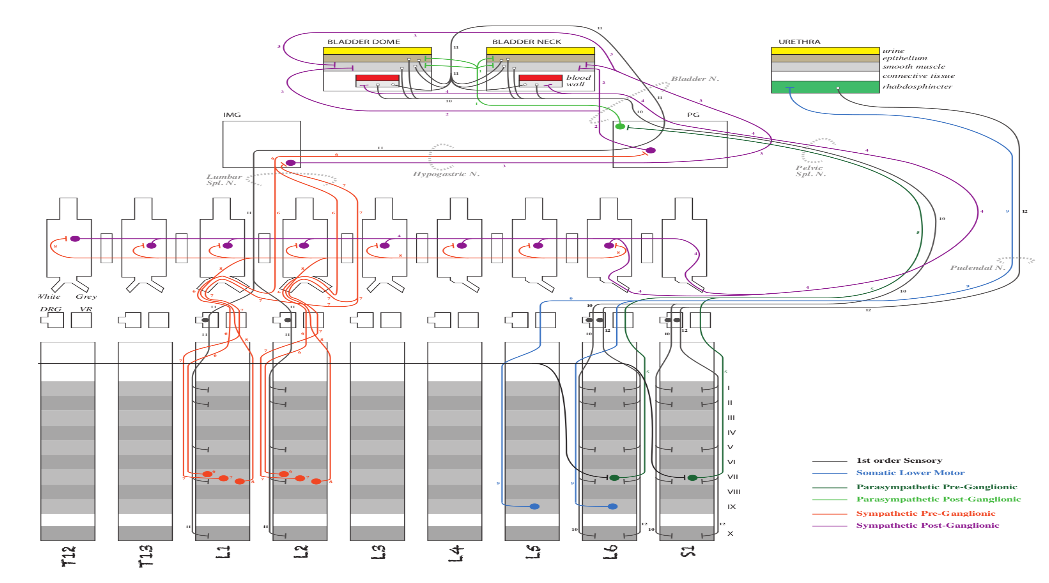}
    \label{fig:keastPart}
}
\subfigure[Scale 3: Neuron modeling] {
    \includegraphics[width=0.47\textwidth]{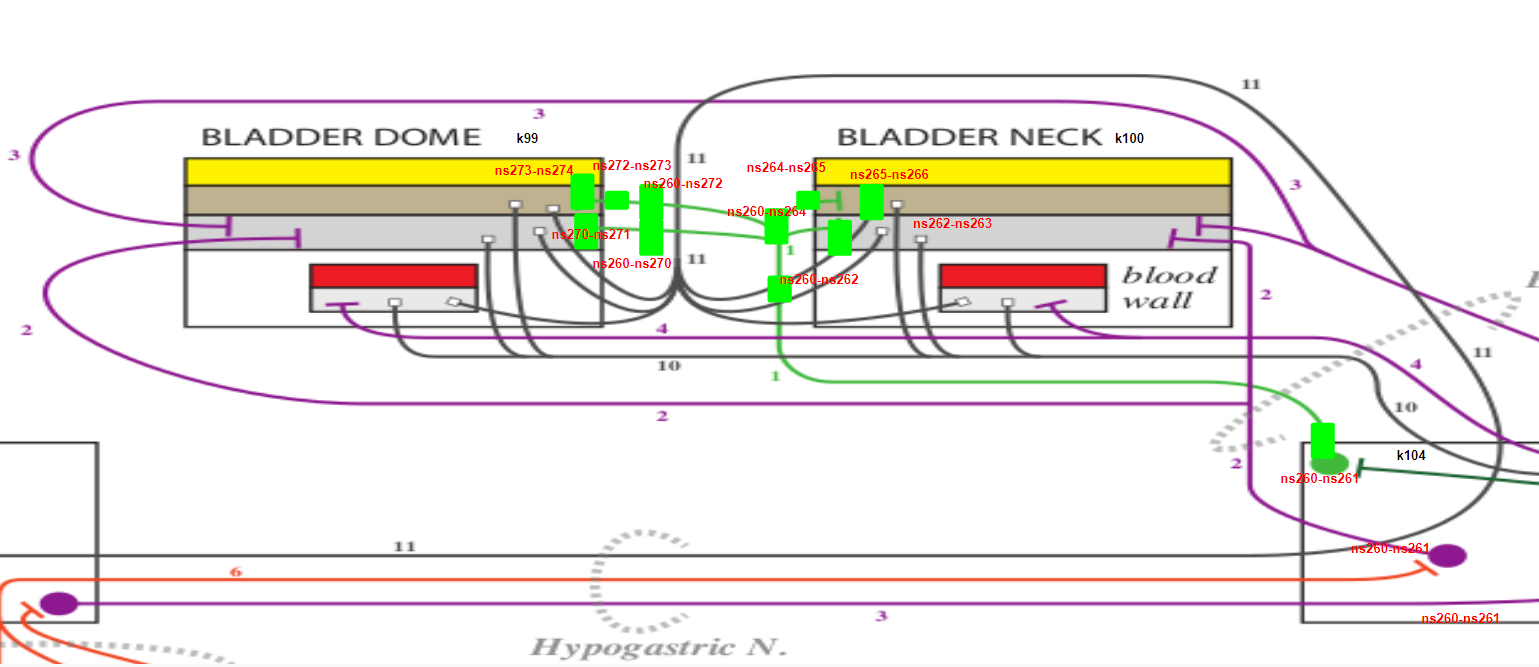}
    \label{fig:keastPart2}
}
\caption{ApiNATOMY model mock-ups}
\label{fig:keast}
\end{figure*}

Taking neural wiring diagrams from the Gray-style representation shown in Figure~\ref{fig:textbook} to the cellular level brings a major increase in complexity. Physiology experts often produce model sketches using general-purpose visual editors. Typically, several drafts are produced, as it is not feasible to capture all aspects in one view. For example, Figure~\ref{fig:keast} shows extracts on three different scales: (i) an overall bladder model of a rat (Figure~\ref{fig:keastFull}), (ii) a higher resolution view of its spinal foldout (Figure~\ref{fig:keastPart}), and (iii) a detailed layout of a particular neuron (Figure~\ref{fig:keastPart2}). Such images can be used for communication, publication, or teaching, but are not suitable for automated integration, analysis, and discovery. The ApiNATOMY modeling framework was designed to provide a data representation format that addresses this issue.  

The ApiNATOMY KR has its basis in the constraints of forming the tissue domain architecture. In particular, experts in physiological systems identify compartmental models of primary Functional Tissue Units (FTUs)~\cite{BSG+15} represented by an abstraction that we call a \emph{lyph}. An overview of ApiNATOMY modeling classes and relations can be found in our prior publication~\cite{kokash2021knowledge}. 
In a nutshell, the model is a graph in which the edges (\emph{links}) are associated with anatomical conduits defined by tissue structures (\emph{lyphs} and \emph{materials}), and the nodes represent points where the structures change or bifurcate. A link can also represent a border of a lyph. Several modeling templates are supported to enable users to define repeated constructs in a simple way. In particular, \emph{chains} help to define the linear concatenation of adjacent segments. In the most straightforward approach, one would need to specify nodes and edges to define a full graph. ApiNATOMY provides multiple templates that require only essential data from the expert user and automatically generates other resources. For example, a chain can be defined by 
\begin{enumerate}
    \item specifying a number of segments (levels) and providing a \emph{lyph template} that gets instantiated into numerous lyphs conveyed by each generated edge;
    \item defining a sequence of lyph instances that get combined into a path in a graph; 
    \item defining a sequence of housing lyphs along which a chain of vascular or neuronal connections passes through. 
\end{enumerate}

All ApiNATOMY modeling classes are related to each other via a vast set of bidirectional relationships, e.g.,
\begin{itemize}
    \item \emph{layers} and \emph{layerIn} connect a lyph with its layers;   
    \item \emph{conveyingLyph} and \emph{conveys} connect a graph link with its lyph definition;
    \item \emph{internalNodes} and \emph{internalIn} position internal nodes on a lyph;
    \item \emph{internalLyphs} and \emph{internalIn} position internal parts of a lyph on its shape; 
    \item \emph{hostedNodes} and \emph{hostedBy} relate a link with nodes positioned on it.
    \item \emph{source} and \emph{sourceOf}, as well as \emph{target} and \emph{targetOf}, relate the links to the node definitions. 
    \item \emph{root} and \emph{rootOf}, as well as \emph{leaf} and \emph{leafOf}, relate the chain ends to the node definitions.  
\end{itemize}
The full list of relationships for each modeling class is available via the ApiNATOMY JSON schema~\cite{ApiNATOMY-tool}. Having all resources described in one specification is helpful at many levels. First, the schema can be used to generate a reference documentation that provides details about the programming elements associated with the modeling framework for modelers or developers looking to adopt it. Second, the JSON specification is used for the syntactic and structural validation of ApiNATOMY input models. It also helps to instantly supply unit tests for various converters we design for user convenience, e.g., spreadsheet XLSX files to JSON files converter, ApiNATOMY specification generators from Neo4J graphs and SQL tables.     
Third, it governs the Javascript-based implementation of the model viewer and editors: the hierarchy and structure of classes and their properties replicate the structure of JSON types and objects representing ApiNATOMY concepts and allows us to easily create object instances from such objects. The classes are supplied with extra methods to handle resources of a given type, e.g., all objects of type \emph{VisualResource} are represented by object instances supplied with \emph{create} and \emph{update} methods that assign visual resources, more specifically, WebGL objects, to visualize model elements such as nodes, links, and lyphs. Finally, keeping internal data representation close to its input format simplifies the serialization of expanded models as we can store generated objects in the same format as the input objects. The serialized generated model also should comply to the schema and can be reloaded to the viewer bypassing the generation step. 

\begin{figure*}
    \centering
    \includegraphics[width=0.97\textwidth]{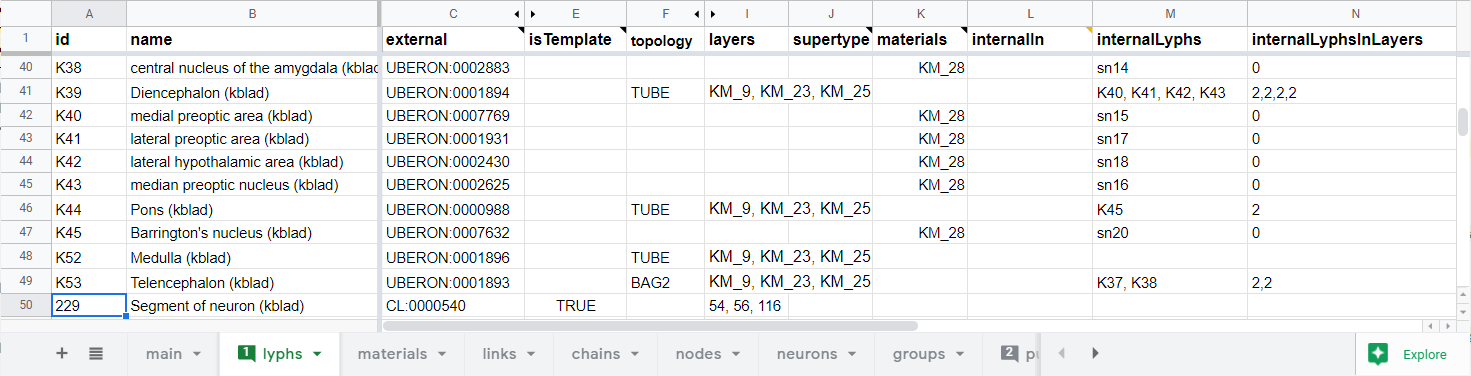}
    \caption{ApiNATOMY input model in spreadsheets}
    \label{fig:keastExcel}
\end{figure*}

After identifying elements of a model, the expert should describe them in terms of ApiNATOMY entities and relationships among them. Technically, model specification can be defined in JSON file, or, in a simpler manner, using ApiNATOMY spreadsheet templates. In such a template, each page is dedicated to the definition of resources of particular type: nodes, links, lyphs, materials, chains, groups, and coalescences. Spreadsheet templates do not support nested definitions, but otherwise allow modellers to define all necessary parameters. Figure~\ref{fig:keastExcel} shows definition of some lyphs for the rat bladder model sketched in Figure~\ref{fig:keast}.   
The ApiNATOMY KM toolset can import such spreadsheets and generate corresponding JSON specifications, and vice versa, a JSON model can be saved as a spreadsheet.

\section{Visual Knowledge Management Tools}
\label{sect:tools} 

\begin{figure*}
    \centering
    \includegraphics[width=0.97\textwidth]{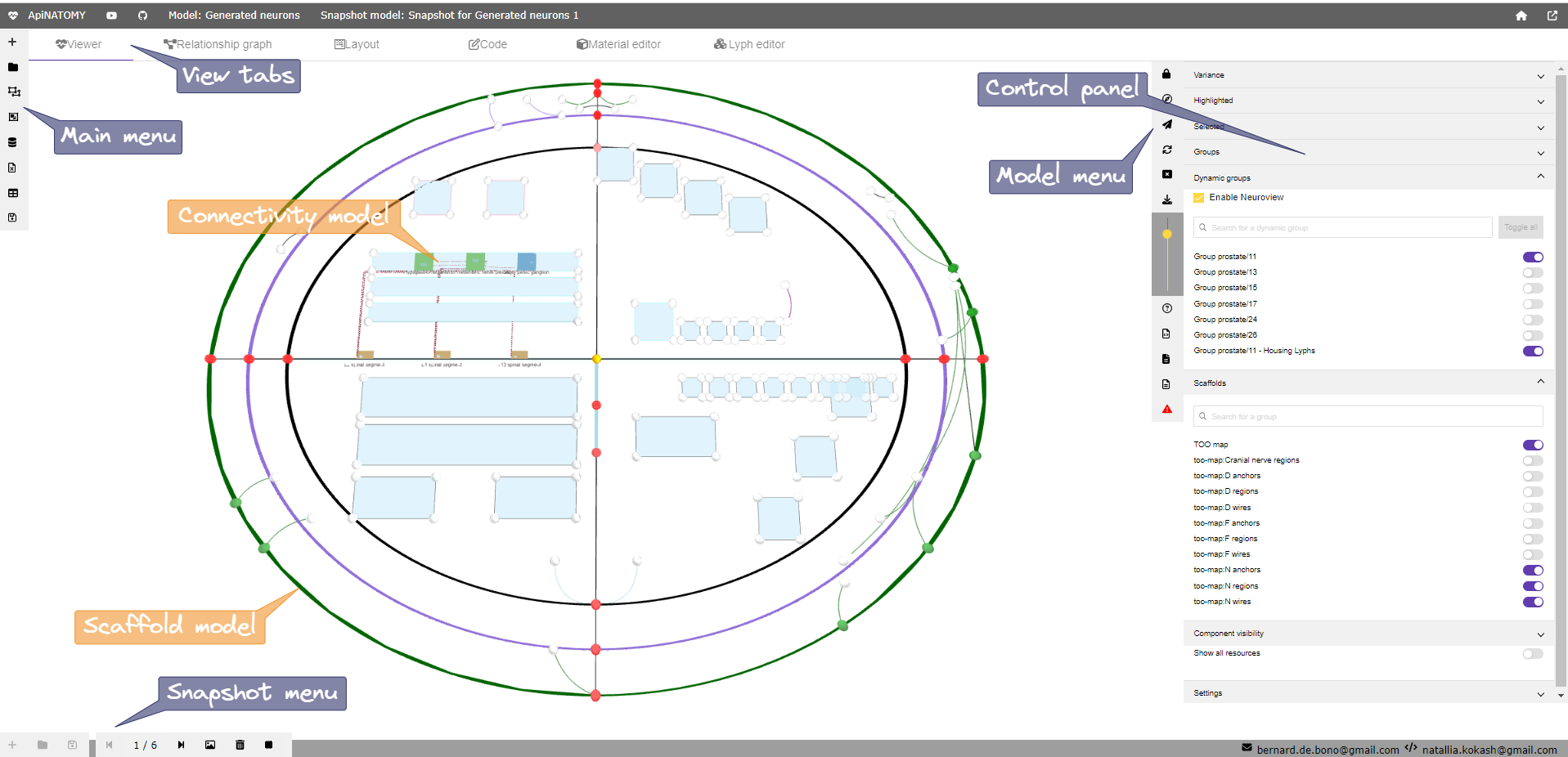}
    \caption{ApiNATOMY visual knowledge management tools}
    \label{fig:overview}
\end{figure*}

The ApiNATOMY model viewer application, shown in Figure~\ref{fig:overview} consists of the following major components: 
\begin{itemize}
    \item \emph{Model viewer} that features a dynamic WebGL graph rendered using a 2D or a 3D force-directed layout algorithm~\cite{kobourov2013force}.
    \item \emph{Control panel} that allows users to change parameters of the viewer and select parts of the model to display.
    \item \emph{Relationship graph} helper component that shows selected relationships among key model resources. This viewer operates on the generated model and hence can be used to inspect derived (auto-generated) resources.
    \item Resource editors
       \begin{itemize}
    \item \emph{Code editor} is a component that shows code of the currently opened ApiNATOMY       JSON specification. It is the most flexible editing tool but requires technical         understanding of the ApiNATOMY schema and model specification conventions.
    \item \emph{Layout editor} allows users to associate physiology model resources with scaffold resources to specify their position within larger body regions.
    \item \emph{Material editor} is a GUI for defining chemical compounds and basic tissue elements used throughout
physiology models.
    \item \emph{Lyph editor} is a GUI for defining key structural resources in ApiNATOMY, lyphs, which are layered compartments composed of materials or other lyphs and represent biological organs or systems.
    \item \emph{Chain editor} is a GUI for defining templates that get expanded to generate chains, i.e., sequences of model graph links conveying lyphs.
    \item \emph{Coalescence editor} is a GUI editor for defining pairs (sometimes, sets) of lyphs with overlapping layers that enable exchange of fluid materials.
    \end{itemize}
    \item Toolbars
     \begin{itemize}
        \item \emph{Main toolbar} allows users to create, load, compose and export data models from the local file system, online repository or a given URL.
       \item \emph{Model toolbar} provides controls for the current graphical scene. It allows users to freeze camera position, reset it to the initial position, toggle control panel, and adjust label font. There are also controls to import external models, export generated model and resource map for integration with the SPARC KB.
       \item \emph{Snapshot toolbar} allows users to save particular scenes. A scene consists of a number of visible groups in the current model, camera position and other parameter settings.
       \end{itemize}
\end{itemize}

Below we present key points about the aforementioned tools.

\subsection{Control panel}

The control panel contains a number of sections with information and GUI controls split into several named accordion panels. The first panel, \emph{Variance}, shows model variations for selected clades. A \emph{clade} is a group of organisms that includes a single ancestor and all of its descendants. This panel is context-dependent and only appears when a model has variances. For example, this concept allows us to reflect differences in the composition of a spinal cord of humans and rodents in the same model. 

The second panel, \emph{Highlighted}, shows conceptual information about visual resources, such as nodes, links and lyphs, as well as equivalent scaffold model resources - vertices, wires, and regions. A highlighted resources gets emphasized with color, and an overlay label with identifier and name appears next to it. The panel shows other important properties of the resource. Similarly, the third panel, \emph{Selected}, shows the information about a visual resource in a more permanent way: it stays in place unless a user selects another resource or clicks on an empty space to reset the panel.

The next panel, \emph{Groups}, shows all groups in the model which a user can include or exclude from the graphical view. A \emph{group} is an important concept that helps us to split the model into logical parts that can be explored separately. The same resource can be included to several groups, and sometimes it is important to show resources that are not explicitly included to the group. For example, if a group includes nodes which are link ends, the link must be shown as well. Generally, whenever a user defines a chain, a groups with all resources constituting this chain is created. 

The \emph{Dynamic groups} panel shows groups which are automatically generated by: 
     \begin{itemize}
        \item a so-called \emph{Neurulator} algorithm,
        \item a SciGraph query handling method.
       \end{itemize}
       
The Neurulator algorithm runs on a new model load and searches for closed components, i.e., subgraph components encompassed by the lyphs with closed borders on all its ends.
Neurulated groups are used to ensure that the individual segments and parts of neurons modeled in ApiNATOMY can be recognized as single cellular entities. By default, ApiNATOMY treats parts of neurons individually so that it is possible to talk about the specific location of a neurite and give it an exact anatomical location. 

The query handling method maps SciGraph resources fulfilling Cypher queries like e.g., \emph{``List all the neuronal processes for a soma located in [start-id] into ApiNATOMY model identifiers''} and encompasses all such resources into corresponding dynamic groups. 
This is useful for querying connectivity defined by neuron populations.
Neuron populations are sets of neurons that share defining properties to the distinguish them from other similar populations. For example, there may be many populations that have their somata located in the Superior Cervical Ganglion, however they can be differentiated by considering their projection targets, both anatomically and based on their target populations.
Axons and dendrites in the ApiNATOMY representation are collective unions of all the individual members of a population. This means that we do not distinguish between cases where a single neuron branches into multiple collaterals that project to different location and multiple neurons that each project to a different location and all combinations in between.

The last panel, \emph{Settings}, allows users to change visual characteritics of the model, e.g., toggle lyphs and lyph layers or switch between 2D and 3D layout. 

\subsection{Scaffolds}
ApiNATOMY \emph{scaffold} is a model for describing reusable layout templates to constrain position and other visual artifacts of connectivity models. Scaffold models define points, called \emph{anchors}, with fixed or constrained coordinates. Similarly to nodes, anchor positions are defined via \emph{layout} property, which, however, accepts 2d coordinates, or via constraints imposed by relationships such as \emph{hostedBy}. Anchors define \emph{wire} ends \emph{region} borders. A context-dependent accordion panel, \emph{Scaffolds}, appears for connectivity models with integrated scaffolds or when a scaffold model itself is opened in the viewer. 

A node can be bound to a scaffold's anchor via its property \emph{anchoredTo}. To make a chain stretch along a given wire, place the reference to the wire to the chain's property \emph{wiredTo}. By default, a wired chain gets its root anchored to the source of the wire, and its leaf to the target of the wire. The direction can be reversed if a Boolean chain's property \emph{startFromLeaf} is set to true. Note that the intermediate nodes of a wired chain directly follow the trajectory of the wire while the intermediate nodes of a chain with just root and leaf nodes anchored to a wire ends are governed by the forced-directed layout algorithm. Lyphs and groups of lyphs can be placed to the area of a scaffold region via their property \emph{hostedBy}.

\begin{figure}
    \centering
    \includegraphics[width=0.5\textwidth]{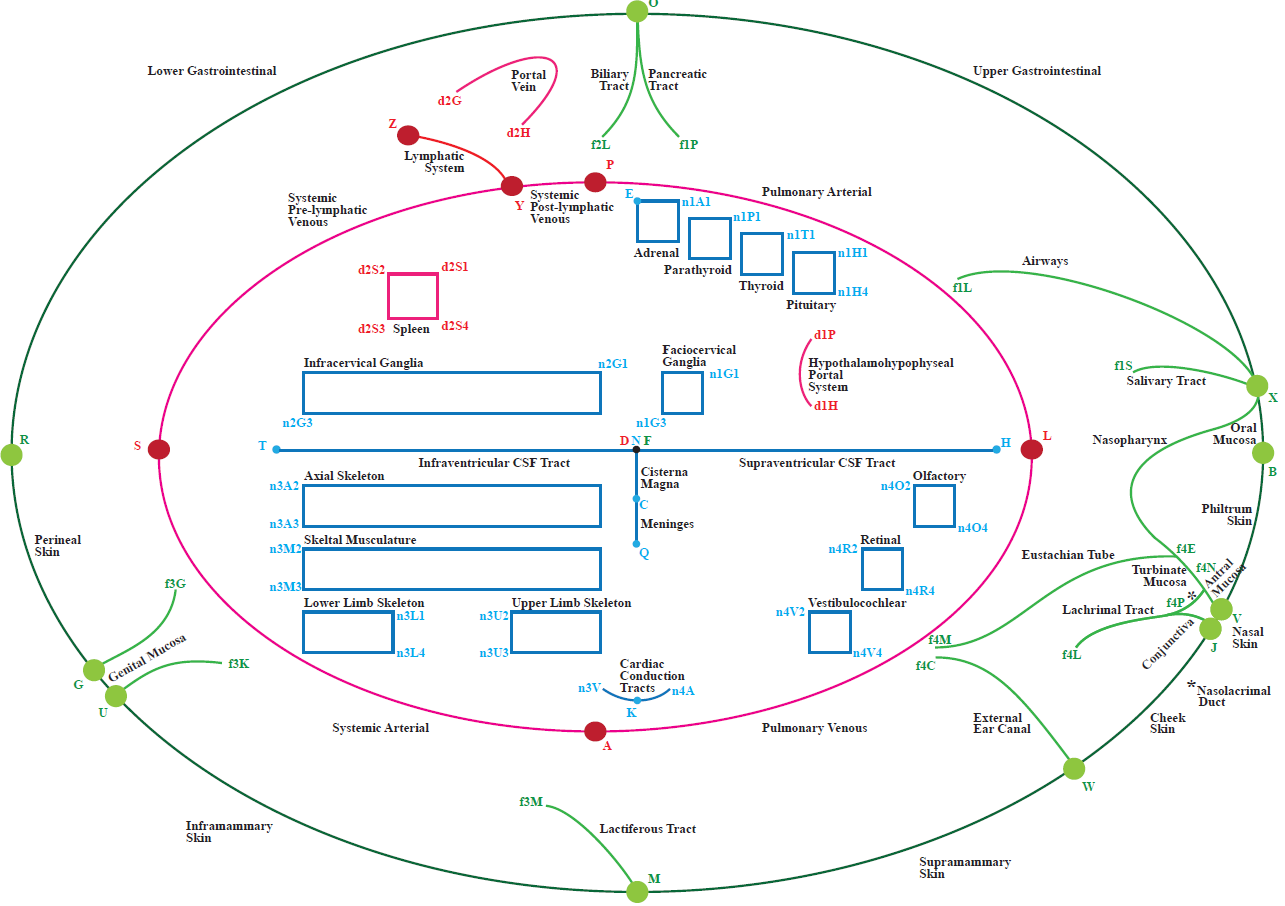}
    \caption{ApiNATOMY scaffold mock-up: Body of a normal vertebrates}
    \label{fig:tooMap}
\end{figure}

The TOO map, shown in Figure~\ref{fig:tooMap} provides the ApiNATOMY author with a wireframe schematic of the body to overview and quality-check connectivity. This subway-style whole-body map depicts as a topological scaffold the three main flow thoroughfares of extracellular material: in blue, a 'T'-shaped depiction of Cerebro-Spinal Fluid (CSF), in red an inner 'O'-shape denotes the circulation of blood, and in green the outer 'O'-shape denotes the flow of materials on the surface of the body, such as digestive juices, food, chyme, chyle, faeces, air, sweat, tears, mucus, urine, milk, reproductive fluids and products of conception. (Hence: 'T' + 'O' + 'O' = TOO). Stylistically revisits a historical technique in map making known as the T-and-O map.

\subsection{Resource editors}
Model editors allow users to define or modify existing ApiNATOMY models using a selected set of predefined GUI-based operations. 

\begin{figure}
    \centering
    \includegraphics[width=0.5\textwidth]{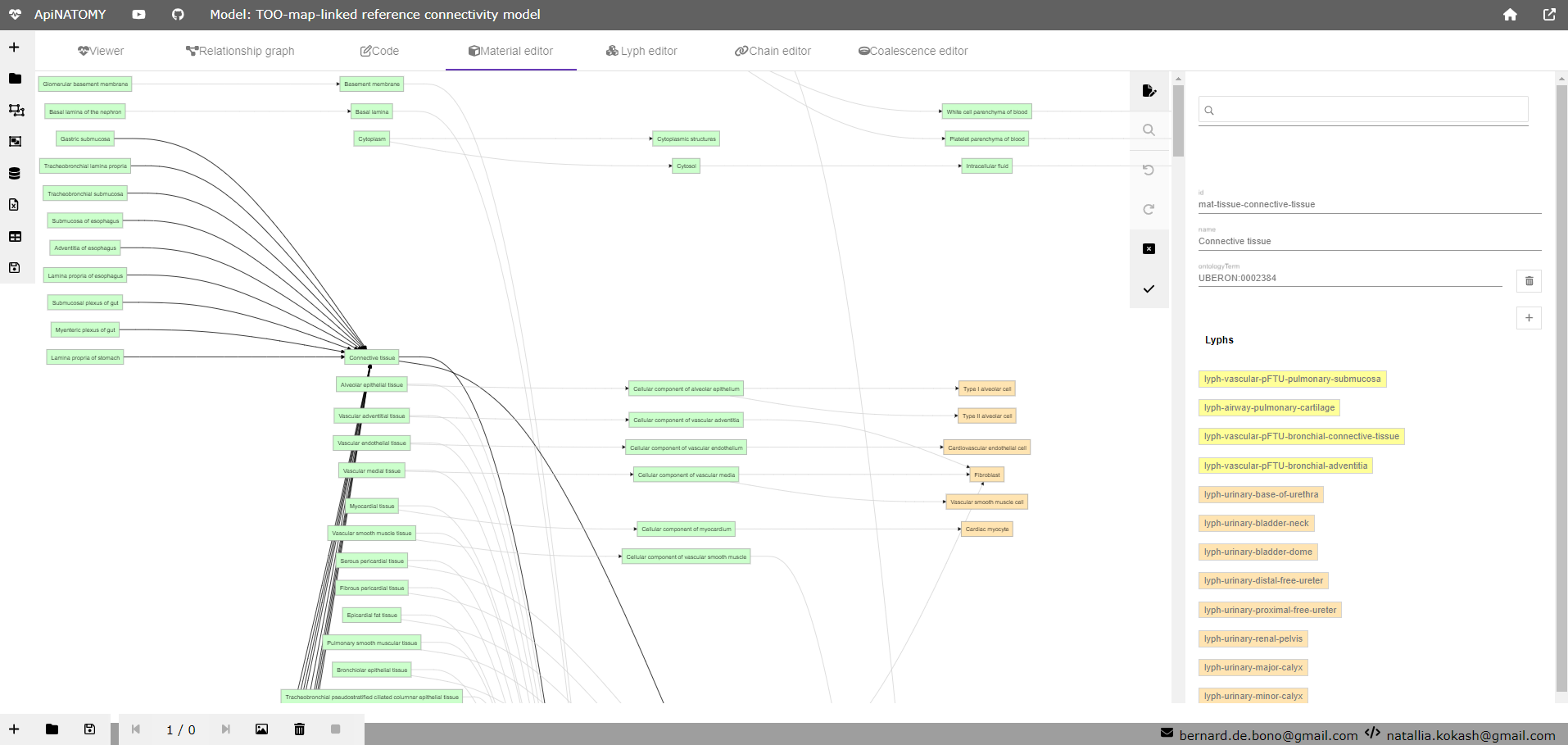}
    \caption{Material editor: DAG view}
    \label{fig:materialEditor}
\end{figure}

\begin{figure}
    \centering
    \includegraphics[width=0.5\textwidth]{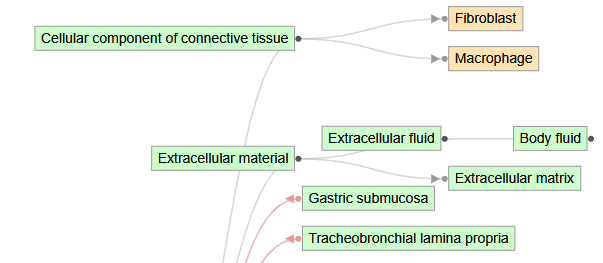}
    \caption{Material editor: a fragment of the tree view}
    \label{fig:materialTree}
\end{figure}

The \emph{Material editor} provides a GUI for defining materials and material-to-material and material-to-lyph relationships. The editor, shown in Figure~\ref{fig:materialEditor} consists of a main view that displays a Directed Acyclic Graph (DAG) of material composition, and a property editing panel that allows users to edit properties of a selected material. The view shows the material composition relationship as defined by the \emph{materials}/\emph{materialIn} properties of materials and lyphs (the latter can be viewed as composite materials). The editor allows users to create, edit and delete materials, add and remove relations to other materials or lyphs, and annotate them with terms from other biomedical ontologies. Properties of the selected material appear in the editing panel, which also shows lyph definitions the selected material is used in as a layer. By clicking on a DAG node, a selected material is highlighted together with its incoming and outgoing edges (see Figure~\ref{fig:materialTree}).

\begin{figure*}
    \centering
    \includegraphics[width=0.97\textwidth]{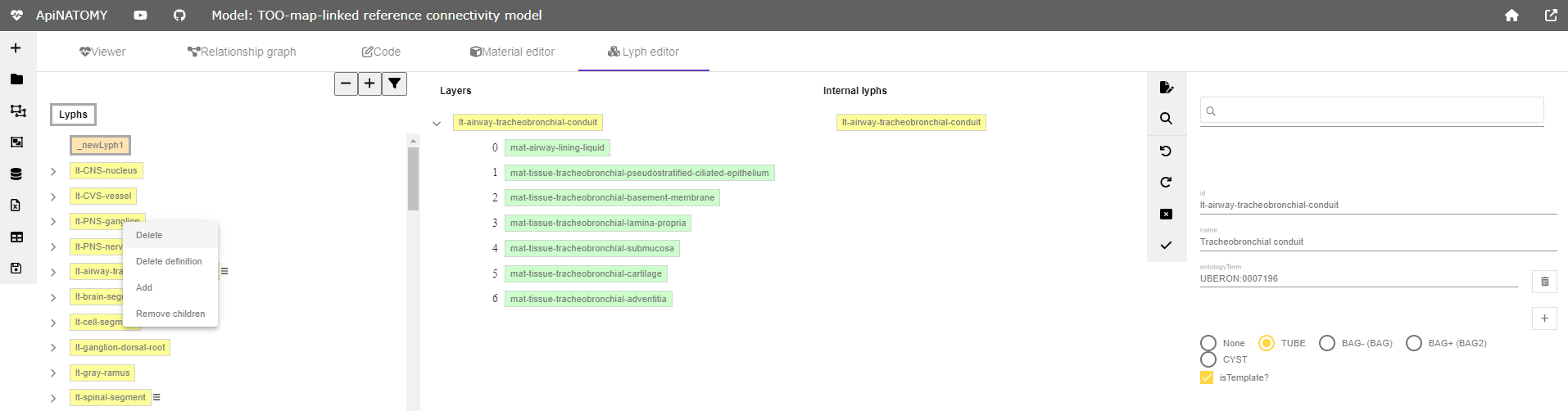}
    \caption{Lyph editor}
    \label{fig:lyphEditor}
\end{figure*}

The \emph{Lyph editor} tab provides a GUI for defining lyphs and lyph-to-lyph relationships. The editor consists of 3 hierarchical trees, and a property editing panel (see Figure~\ref{fig:lyphEditor}). The first tree view shows the list of declared lyphs arranged into hierarchy by the \emph{sypertype}/\emph{subtypes} relationships. Lyphs and lyph templates in this view can be distinguished by the tile color. By clicking on the lyph identifier in this component, one selects a lyph to inspect and modify. The second view shows the layers of the selected lyph defined by the \emph{layers}/\emph{layerIn} relationships. The third view shows internal lyphs of the selected lyph defined by the \emph{internalLyphs}/\emph{internalIn} relationships. Users can identify lyphs that have layers and/or internal lyphs by presence of special icons next to the lyph name in the lyph view.

Create, Read, Update, and Delete (CRUD) operations in the editor are context-dependent - they are applied to a selected lyph in an active tree and vary depending what tree is active. Each applied operation is registered and can be reverted or repeated using ``Undo'' and ``Redo'' buttons. 

\begin{figure}
    \centering
    \includegraphics[width=0.5\textwidth]{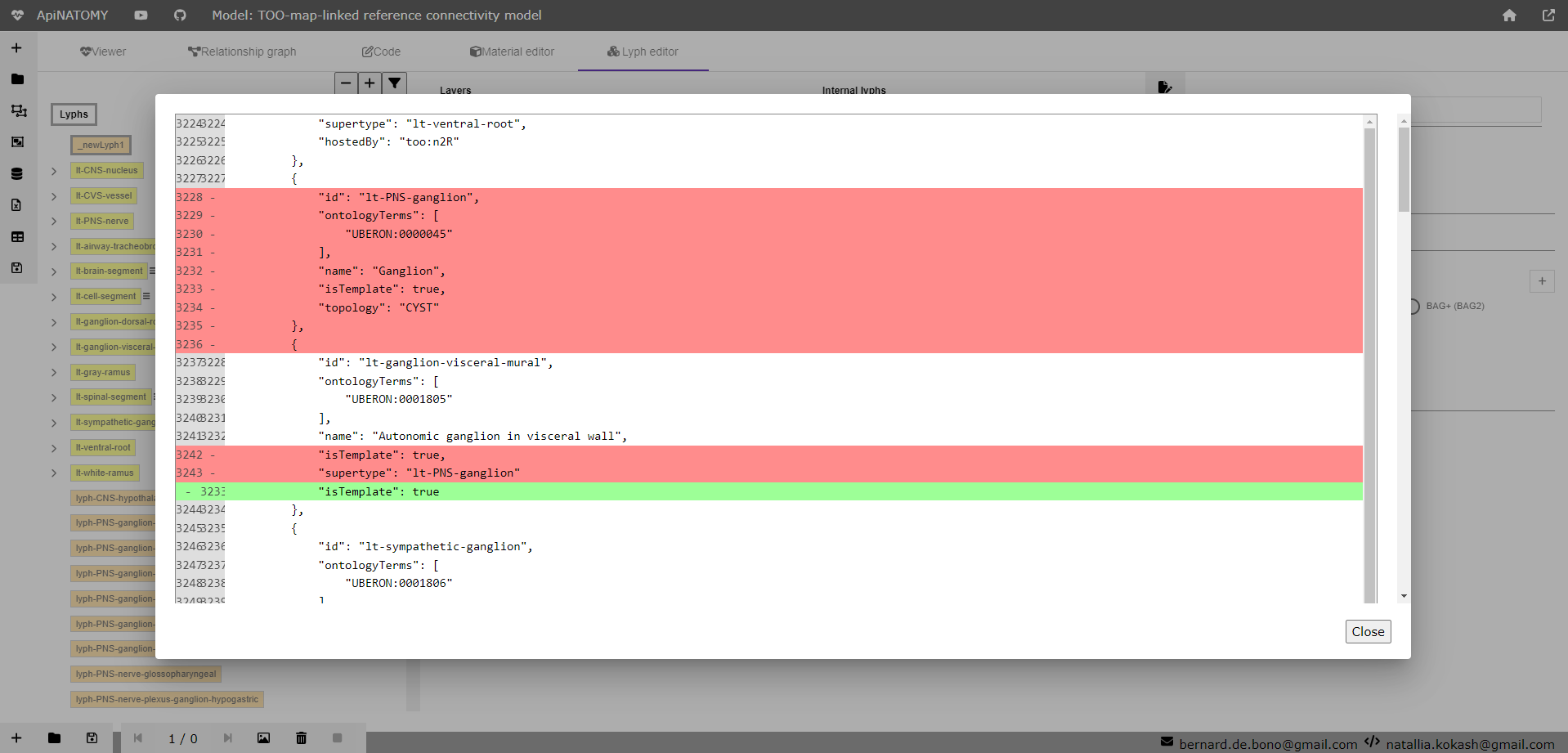}
    \caption{Lyph editor: review changes}
    \label{fig:lyphEditor-compare}
\end{figure}

The operations performed with the help of the GUI are translated into JSON code. The revised model code can be compared with previous versions using a code comparison tool. For example, Figure~\ref{fig:lyphEditor-compare} illustrates model changes after a lyph template named ``Ganglion'' was deleted. This lyph was used in the definition of a lyph named ``Autonomic ganglion in the visceral wall'', so this definition was revised as well. 

Other resource editors, \emph{Chain editor} and \emph{Coalescence editor} provide similar style GUIs for managing chain and coalescence definitions, respectively. Apart from CRUD operations, the chain editor supports operations to manipulate the connectivity graph at a larger scale, e.g., split, merge, or clone subgraphs. Lateralization in physiology refers to the functional specialization of the left and right sides of the body and is often overlooked by modeling tools. With the help of the ApiNATOMY editors, modelers can quickly replicate the existing non-lateralized structure and then further edit it to reflect the differences in left or right part of the organ.

The property panels in all editors provide fields to specify resource properties such as \emph{id} and \emph{name}, but are also integrated with the SciGraph querying API, allowing users to search for relevant terms in other ontologies and include them in the model definitions via the \emph{ontologyTerms} property. Such annotations are essential for automatically merging ApiNATOMY models with other SCKAN resources.

The editors are equipped with validation and error handling services to prevent users from creating inconsistent models. Operations such as deletion of a resource or changing its identifier trigger the entire model revision to clear or update references to the modified resource.  

\subsection{Combining models}

Once a model is created, it can be stored locally or in a shared model repository as a JSON file or a spreadsheet. For example, SPARC ApiNATOMY models in both formats are available here~\cite{ApiNATOMY-models}. 

Experts can reuse and combine existing models to build a larger model of a physiological system. There are several ways to integrate several models into one:
  \begin{itemize}
    \item Merge models: load a base model into the viewer from a file or an online repository, and, using the menu button, \emph{merge} another model into the current one. With this method, resource definitions are mixed together.     
    \item Join models: load a base model into the viewer from a file or an online repository, and, using the menu button, \emph{join} another model into the current one. With this method, models are kept separate in new groups within the joi model.   
    \item Import models: models published online can be included into a new model via the \emph{imports} statement. The viewer can download the imported models via HTTP requests to the specified locations. The definitions from the imported models are used to instantiate the visual layout and get exported as part of the generated model for integration with external knowledge bases such as SciCrunch~\cite{SciCrunch}, but they are not included into initial specification and must be downloaded every time a user opens the model in the viewer.   
    \item Add scaffold: scaffolds are models that help arranging connectivity specifications. They can be added to a model using one of the above methods. If at least one scaffold is present in the model, he \emph{Layout editor} becomes available, in which users can define associations between connectivity model concepts and the scaffold concepts.  
  \end{itemize}

If there are resources with the same identifier, the combined model may have duplicate definitions, and a corresponding warning will be issued. To prevent such problems while combining models, we support \emph{namespaces}, abstract environments to hold logical groupings of unique identifiers or symbols. 
If a certain resource is referenced in a model but not defined explicitly, it is automatically generated, unless this reference includes a prefix pointing to another namespace. 

\begin{figure*}
    \centering
    \includegraphics[width=0.97\textwidth]{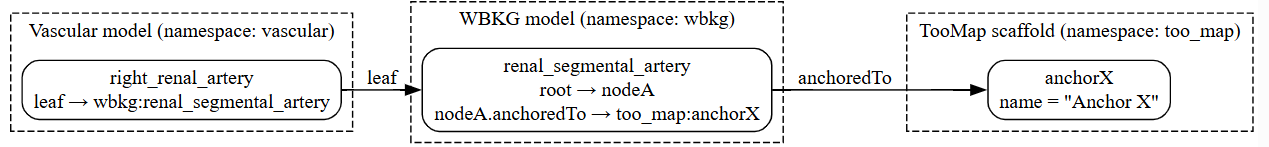}
    \caption{Multi-model assembly: Connecting Left Renal Artery (LRA) chain from the Vascular model to the root of the  Renal Segment Artery (RSA) chain from the WBKG model. The root of the RSA is forced to a position defined by the anchor defined in the TooMap scaffold.} 
    \label{fig:multi-namespace}
\end{figure*}

A model with references to resources from other namespaces can be visualized in the editor, but the validation module marks it as invalid until the model with necessary definitions is added. Models with errors, including missing resources, are marked with a yellow (warnings) or red (critical errors) icon. By clicking on this icon, users can overview logged messages that indicate what errors were detected in the model by the validation module. 
         
\subsection{Layout rendering}     
The visualization displays graph $G = \{V, E\},$ where $V$ is a set of visible nodes, and $E$ is a set of visible links. Nodes and links are visible if their property \emph{isVisible} is set to true. This property is assigned to nodes and links in a group when a user toggles groups in the \emph{Settings panel}. By default, visible nodes will be placed into positions determined by the force-directed layout algorithm at each iteration. However, node position can be controlled in several ways:
\begin{enumerate}
    \item Node position coordinates are given in its \emph{layout} property. The coordinates provided in this property create an attraction point (magnet) for this node in the force-directed layout. If the node's property \emph{fixed} is set, the node is firmly positioned in that point. 
    \item A node position can be determined as the center of coordinates of its \emph{controlNodes} set. 
    \item A node position can be determined by its hosting link. We say that a link hosts a node if the node refers to the link via its \emph{hostedBy} property. Optionally, a node's \emph{offset} property can indicate the fraction of the distance of the node from the link's source end along the link's curve. Since most of the relations in the ApiNATOMY schema are bidirectional, the link is aware about its hosted nodes via the \emph{hostedNodes} property. Nodes without given offset are positioned on equal distance from each other along the link's curve, i.e., their offset is computed as $1/|hostedNodes|,$ where $|.|$ indicates the set dimension.
    \item The nodes positioned on a lyph's border are implicitly positioned on an invisible link coinciding with the border. 
    \item A node can be placed into the center of a lyph if its property \emph{internalIn} points to this lyph. If a lyph is associated with several nodes via its \emph{internalNodes} property, these nodes are placed in a circular or a grid pattern within the lyph's shape.
    \item Nodes can be bound to scaffold anchors via their \emph{anchoredTo} property. When present, anchoring overrides other positioning methods. 
\end{enumerate}

Link positions are defined by the positions of their end nodes, and the lyphs are aligned along the links they convey. Lyphs can freely rotate around the link they convey. Usually, the rotation angle is set to 0, but for lyphs involved into coalescences, the rotation angle is selected such that coalescing lyphs can share a common layer. The presence of connecting coalescences in the model also affects the positioning of nodes as coalescing lyphs must be placed on parallel lines next to each other.  

\begin{figure}
    \centering
    \includegraphics[width=0.5\textwidth]{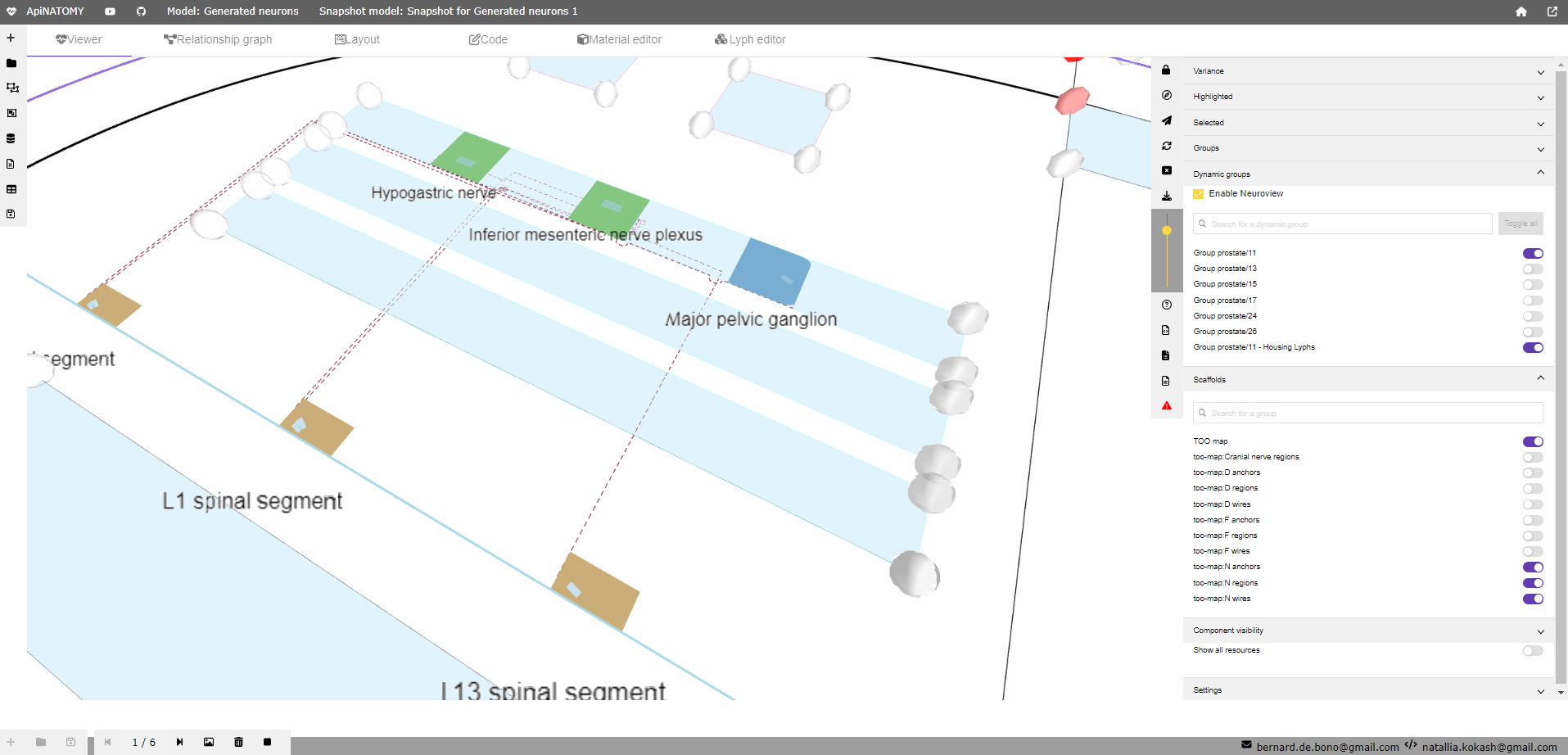}
    \caption{ApiNATOMY viewer: automatically generated layout for a neuron group} 
    \label{fig:overview-zoomed}
\end{figure}

Links that are wired to scaffold wires are stretched along these wires, and lyphs hosted by scaffold regions are positioned within the region. If lyphs within a region or another lyph are combined into a chain, a method minimizing edge crossing is applied. Figure~\ref{fig:overview-zoomed} shows an example of a neuron chain allocation along hosted lyphs with the use of the edge crossing minimization. 

ApiNATOMY targets multi-scale modeling. That means that large elements such as body parts may appear next to tiny elements such as neuron chains and cells, in the same model. The modeling notation allows users to specify both representational dimensions for the visualization and anatomical dimensions for educational purposes, scientific analysis and/or automated reasoning. 
However, researches concerned with accurate topological and biomedical modeling of an anatomical system may not be willing to focus immediately on the relative scaling of involved resources. Therefore, we provide default scaling rules that resize lyphs based on constrains implied by their relations to other elements of the model. For example, internal lyphs of a lyph or a region must all be scaled to fit into the hosting lyph or the region. The size of a  lyph conveyed by a link depends on the length of the link, while its proportions depend on the number of the layers in this lyph.        

\section{ApiNATOMY as part of SCKAN}
\label{sect:infrastructure}

\begin{figure*}
    \centering
    \includegraphics[width=0.97\textwidth]{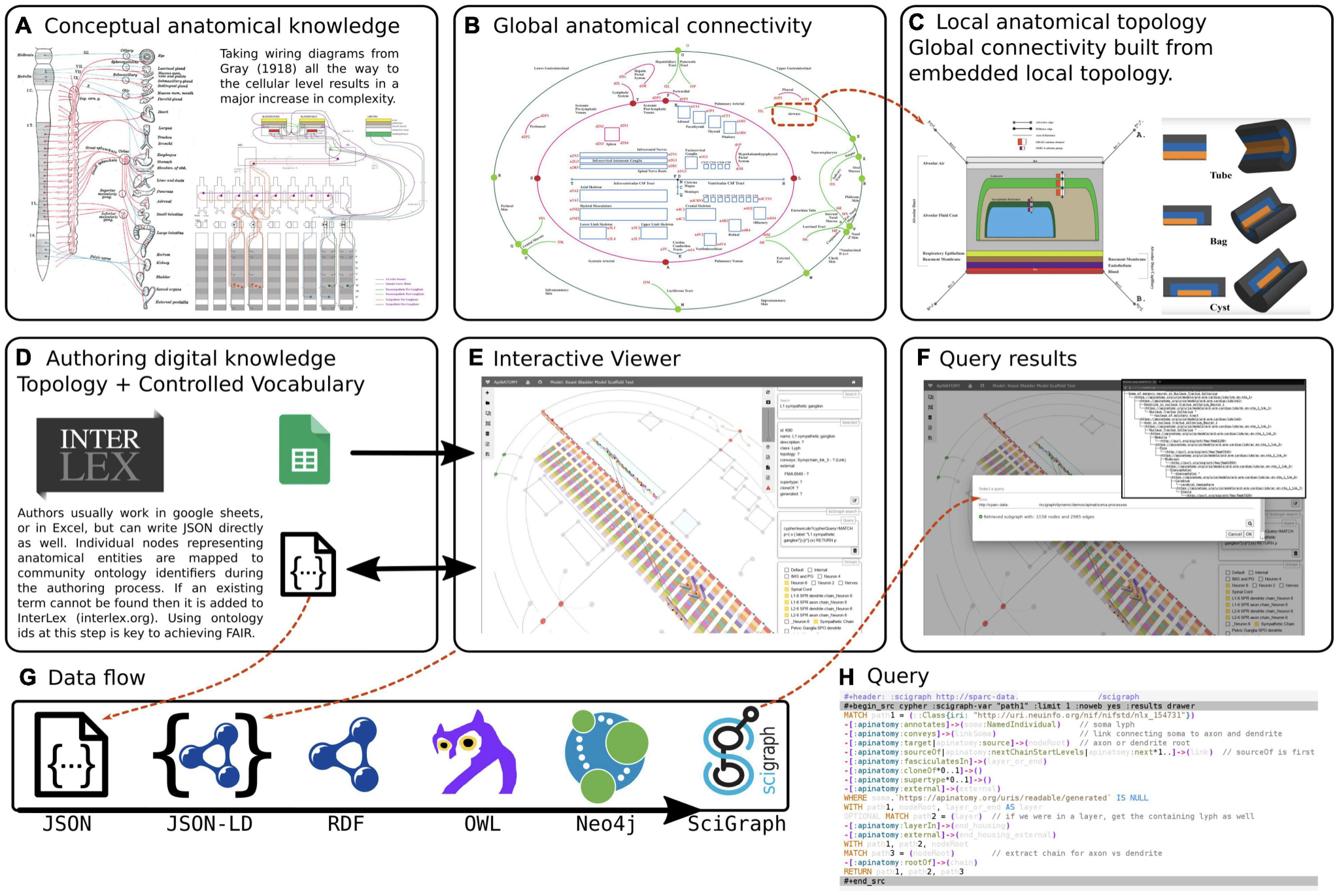}
    \caption{Integration of ApiNATOMY models to the SPARC infrastructure}
    \label{fig:spark-infrastructure}
\end{figure*}

The workflow that brings ApiNATOMY models into SCKAN is shown in Figure~\ref{fig:spark-infrastructure}. The models designed using our tools, after validation, template expansion and resource generation, are exported as JSON-LD files. From JSON-LD, the data gets translated to the RDF/OWL and Neo4J to become part of the SciGraph. SCKAN has a structured semantic form and provides endpoints for accessing and searching its knowledge.    

The visual toolset presented above supplies mechanisms for biomedical experts to compose and validate multi-scale anatomical models. The automatically generated layouts are helping experts in curating data, but they significantly differ from traditional depiction of anatomy and connectivity circuits and may not be clear to experts unfamiliar with the ApiNATOMY toolkit. The SPARC Portal has another way to inspect connectivity data.    

Flatmaps~\cite{flatmaps} provide 2D graphical representations of the anatomy, functionality, and topology of the connectivity of the peripheral nervous system, in a Google Maps-like web-interface. There are two types of flatmaps: an Anatomical Connectivity (AC) flatmap and a Functional Connectivity (FC) flatmap. The AC flatmaps provide a simple graphical representation of the anatomical structures of different species, including human, rat, mouse, cat, and pig, with overlays of nerve connections. Each flatmap consists of two main regions: the body including the visceral organs, and the CNS with more detail of the brain and spinal cord. These two regions are connected through directional edges to demonstrate neural connectivity between the brainstem and the organs of the body.  

The flatmaps consist of several distinct components:
\begin{itemize}
    \item The base anatomical diagrams: the manually-drawn anatomical cartoons, which form the base of each species’ flatmap, annotated with anatomical identifiers corresponding to those used in SCKAN. The base diagram for the FC flatmaps is a hand-drawn map of blocks representing 12 organ systems, including all organs and FTUs within each organ. All entities are annotated with anatomical identifiers consistent with SCKAN.
    \item The connectivity knowledge: captured via ApiNATOMY knowledge models and other knowledge sources and encoded as consistent connectivity knowledge in SCKAN. In AC flatmaps, the connectivity pathways are at the level of neural sheaths, while in FC flatmaps, the pathways reflects single neuron connections.
    \item The mapmaker: a tool which combines the annotated base diagrams with the connectivity knowledge to automatically render the connectivity into map tiles to be used in presenting the flatmaps.
    \item The flatmap viewer application: the JavaScript application to render flatmaps obtained from a server and allow a user to interactively explore them.
    \item FlatmapVuer: the widget used on the SPARC Portal that wraps the flatmap viewer to provide SPARC specific functionality.
    \item The flatmap server: a web server that contains generated flatmaps, both map tiles and metadata about the flatmaps.
\end{itemize}

\begin{figure}
    \centering
    \includegraphics[width=0.5\textwidth]{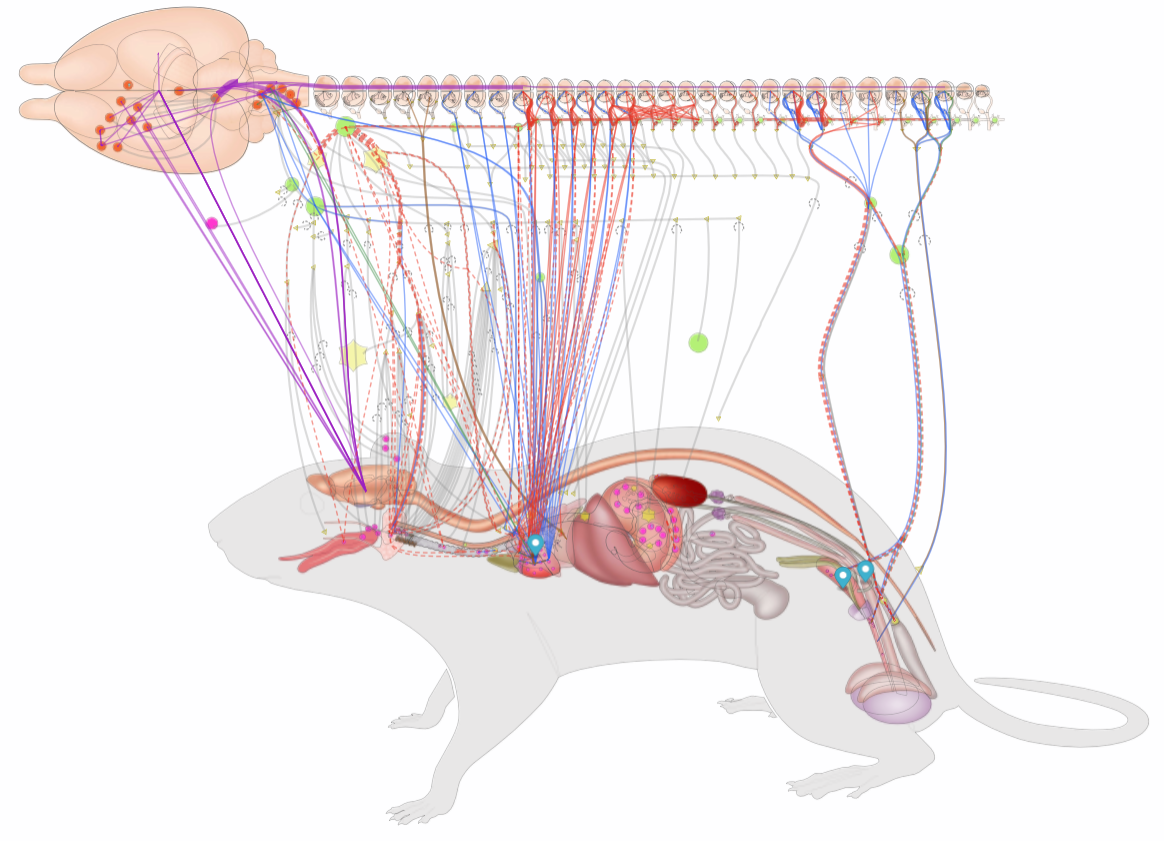}
    \caption{Flatmap visualization of the ApiNATOMY neuron models}
    \label{fig:rat-map}
\end{figure}

Figure~\ref{fig:rat-map} shows a sample flatmap with neural connectivity data in rodents supplied via the ApiNATOMY modeling. More information about the production of animated flatmaps from the ApiNATOMY models of connectivity can be found at the SPARC Portal~\cite{apinatomyMap}.

\section{Related Work}
\label{sect:relatedWork}

Biomedical ontologies are structured frameworks that facilitate the organization, integration, and analysis of biological and medical data~\cite{bioOnto}. These ontologies enable standardized representations of knowledge, which is essential for data sharing, computational analysis, and improving decision-making in clinical and research settings. Since 2000, the field has witnessed significant growth, marked by the development of numerous ontologies, methodological advancements, and increasing integration into various biomedical applications.

The early 2000s saw the emergence of foundational biomedical ontologies that aimed to address the growing need for standardized vocabulary in biomedicine. One of the most influential works during this period was the creation of the Gene Ontology (GO)~\cite{ashburner2000gene,GO}. GO provides a structured vocabulary for annotating genes and gene products based on three main aspects: molecular function, biological processes, and cellular components. This ontology was pivotal in standardizing genetic data and remains widely used in genomics research.

In parallel, the Foundational Model of Anatomy (FMA) ontology was developed to provide a comprehensive model of the human anatomy~\cite{FMA}. The FMA was one of the first ontologies to emphasize a formal structure, enabling computational reasoning about anatomical entities and their relationships. 

Another critical contribution during this era was the Open Biomedical Ontologies (OBO) Foundry, a collaborative initiative established in 2007 to create a suite of interoperable reference ontologies in the biomedical domain~\cite{smith2007obo}. The OBO Foundry promotes the development of ontologies that adhere to shared principles, including openness, orthogonality, and a commitment to collaboration, ensuring that the ontologies are compatible and can be used in combination for broader applications.
The Cell Ontology (CL) is an OBO Foundry ontology covering the domain of biological cell types with curation focused on animal cell types and interoperability~\cite{CL}.

The period from 2010 to 2020 was marked by significant methodological advancements in ontology development, curation, and application. This decade saw the growing integration of biomedical ontologies with big data analytics, allowing researchers to harness large-scale datasets from genomics, proteomics, and electronic health records. Ontologies began playing a critical role in clinical decision support systems, enhancing the ability to link patient data with clinical guidelines and research evidence~\cite{bioOnto-formats}. Ontology development tools, such as Protégé, became more sophisticated, enabling researchers to create, manage, and visualize complex ontologies easily~\cite{musen2015protege}. Furthermore, the rise of automated and semi-automated ontology generation techniques, leveraging machine learning and natural language processing, began to accelerate ontology development, making it more scalable and less reliant on manual curation.

One notable advancement was the development of the Human Phenotype Ontology (HPO), designed to provide a standardized vocabulary for describing phenotypic abnormalities in human disease. HPO has been instrumental in clinical genetics, supporting phenotype-driven diagnostics and research~\cite{robinson2008human}. The integration of HPO with other ontologies has significantly enhanced computational disease modeling and the identification of genotype-phenotype correlations. 
Ontologies for precision medicine also gained traction during this period, notably the Disease Ontology (DO) and the Ontology for Biomedical Investigations (OBI)~\cite{schriml2012disease}. Deep phenotyping and precision medicine have further highlighted the importance of ontologies in linking genetic data with clinical outcomes. The Monarch Initiative  uses ontologies like HPO and the Mammalian Phenotype Ontology (MPO) to integrate phenotype data across species, improving our understanding of human diseases through cross-species analysis~\cite{mungall2017monarch}. Chemical Entities of Biological Interest(ChEBI)  is a chemical database and ontology of molecular entities focused on `small' chemical compounds~\cite{CHEBI}. 

The advent of FAIR data principles has also influenced the development and adoption of biomedical ontologies. Ontologies are increasingly seen as essential tools for making biomedical data FAIR-compliant, thus enhancing data sharing and reuse in both research and clinical settings~\cite{wilkinson2016fair}.

From 2020 onwards, the focus on ontology integration, scalability, and real-world applicability intensified. One of the key trends has been the increased use of artificial intelligence (AI) and machine learning (ML) in ontology development and application~\cite{filice2021biomedical}. AI-driven approaches are now employed to mine biomedical literature, automatically update ontologies, and identify new relationships within existing frameworks, thus reducing the time and cost associated with manual ontology curation~\cite{he2023applications}.

Efforts have been made to address the challenges of ontology interoperability and scalability. The Physiome Project is the worldwide effort to define the physiome via databasing and the development of integrated quantitative and descriptive modeling~\cite{HRN02,VH16,RC08}. The NIH Common Fund Human Biomolecular Atlas Program (HuBMAP) is devoted to the development of a widely accessible framework for comprehensively mapping the human body at single-cell resolution~\cite{HuBMAP}.

Multi-scale modeling is an integrative approach that examines biological systems across various spatial and temporal scales, from molecular interactions to whole-organism behaviors~\cite{DM+11,WPP13,EKB+11}. By bridging these scales, researchers develop comprehensive models that reflect the complexity of biological phenomena, such as disease progression and treatment responses~\cite{DWZ+11,WAB+2014}. This approach combines data from sources like genomics and imaging studies, enabling predictions of system-level behaviors and identification of therapeutic targets. Ultimately, multi-scale modeling enhances drug development, informs clinical decision-making, and improves patient outcomes, making it a vital tool in advancing our understanding of health and disease~\cite{CTC15,MGB+15}.

Visualization tools for biomedical ontologies are crucial for understanding complex relationships within large datasets, facilitating ontology development, and enhancing data interpretation. These tools help researchers and clinicians navigate intricate networks of biological concepts, supporting tasks such as ontology curation, validation, and application in research~\cite{ontoVis-survey}.

One of the most widely used tools is Protégé, which offers a comprehensive environment for ontology editing and visualization. Protégé supports visual exploration of ontological hierarchies and relationships, making it a cornerstone tool for ontology developers and domain experts~\cite{musen2015protege}. 
OntoGraf, an extension of Protégé, enhances ontology visualization by providing interactive graphical representations of classes and their relationships, aiding in the identification of ontology structure and inconsistencies.
Cytoscape, originally developed for molecular interaction networks, has also been adapted to visualize ontologies, allowing users to explore large-scale ontological datasets and analyze the connections between different biological entities~\cite{shannon2003cytoscape}. BioPortal offers web-based visualization of ontologies, allowing researchers to browse and compare various biomedical ontologies. This tool supports visual exploration of ontology mappings, enhancing data integration efforts in biomedical research~\cite{noy2009bioportal}.

The SPARC initiative is built on top of many notable ontologies, namely:
\begin{itemize}
    \item EMAPA, an ontology of mouse anatomical terms, represents developmental and postnatal mouse anatomy in a standardized and searchable format~\cite{hayamizu2013mouse}.
    \item FMA, an evolving computer-based knowledge source for biomedical informatics,
    \item GO, the world’s largest source of information on the functions of genes, and a foundation for computational analysis of large-scale molecular biology and genetics experiments in biomedical research.
    \item UBERON, a multi-species anatomy ontology~\cite{mungall2012uberon}.
\end{itemize}
but also offers novel knowledge repositories:
\begin{itemize}
    \item SAWG InterLex, integrated anatomical knowledge maintained by the SPARC Anatomy Working Group (SAWG), which provides vocabulary services, term request and connectivity.
    \item The NIFSTD ontology, the backbone of the SPARC vocabularies~\cite{bug2021nifstd}.
    \item The Neuron Phenotype Ontology (NPO): A FAIR Approach to Proposing and Classifying Neuronal Types~\cite{gillespie2022neuron}.
    \item SciGraph, an open-source Neo4J ontology store that serves the SPARC vocabularies and houses the SPARC KG.
\end{itemize}

Our approach to curating connectivity data for SPARC relies on a specialized methodology with KR and KM tools that prioritise user convenience and transparency. ApiNATOMY models can be defined as spreadsheets, a straightforward format for viewing and exploring data by researchers. Spreadsheet models, automatically converted to JSON files, governed by the ApiNATOMY JSON schema, are then visualized in schematics that preserve spatial constraints implied by the resource relationships. Equipped with validation, annotation, and graphical editing tools, the KM toolset helps users to create accurate multi-scale models of connectivity which can be smoothly integrated with other databases and tools developed by the biomedical research community.         

\section{Conclusions and future work}
\label{sect:conclusions}

Biomedical ontologies have evolved significantly from simple vocabularies to complex frameworks that underpin modern biomedical research and healthcare. Their role in data standardization, integration, and analysis has made them indispensable components of any large-scale innovation initiative. As ontology development continues to leverage AI and advanced data analytics, the impact of these tools on biomedicine is expected to grow even further.

The organizational and technical effort involved into the creation of such extensive knowledge bases often remains behind the scenes. In this manuscript, we elicited the extensive effort that goes into the collaborative assembly of specialized data, which often requires the development of suitable editing and quality validation tools. We presented the KM tools from the ApiNATOMY framework, a modeling methodology for multi-scale physiology systems. We described the model editor and the model components, outlined resource visualization algorithms, and illustrated their use within the SPARC infrastructure.

In our future work, we will continue to evolve the KR and KM tools to tackle challenges of physiological system modeling and enable quantitative multi-scale analysis~\cite{BSG+17}. We also extensively work on creating an educational module  for ApiNATOMY to make more field experts familiar with the modeling methodology it provides.      

\bibliographystyle{SageH}
\bibliography{main}

\begin{thebibliography}{55}
\providecommand{\natexlab}[1]{#1}
\providecommand{\url}[1]{\texttt{#1}}
\providecommand{\urlprefix}{URL }
\expandafter\ifx\csname urlstyle\endcsname\relax
  \providecommand{\doi}[1]{DOI:\discretionary{}{}{}#1}\else
  \providecommand{\doi}{DOI:\discretionary{}{}{}\begingroup
  \urlstyle{rm}\Url}\fi

\bibitem[{Antezana et~al.(2009)Antezana, Kuiper and Mironov}]{bioKM}
Antezana E, Kuiper M and Mironov V (2009) {Biological knowledge management: the
  emerging role of the Semantic Web technologies}.
\newblock \emph{Briefings in Bioinformatics} 10(4): 392--407.
\newblock \doi{10.1093/bib/bbp024}.

\bibitem[{Ashburner et~al.(2000)}]{ashburner2000gene}
Ashburner M et~al. (2000) Gene ontology: tool for the unification of biology.
\newblock \emph{Nature Genetics} 25: 25--29.

\bibitem[{Bodenreider and Stevens(2006)}]{bioOnto-formats}
Bodenreider O and Stevens R (2006) {Bio-ontologies: current trends and future
  directions}.
\newblock \emph{Briefings in Bioinformatics} 7(3): 256--274.
\newblock \doi{10.1093/bib/bbl027}.

\bibitem[{Bug et~al.(2008)Bug, Ascoli, Grethe, Gupta, Fong and
  Martone}]{bug2021nifstd}
Bug WJ, Ascoli GA, Grethe JS, Gupta A, Fong RS and Martone ME (2008) {The
  NIFSTD and BIRNLex vocabularies: building comprehensive ontologies for
  neuroscience}.
\newblock \emph{Neuroinformatics} 6(3): 175--194.
\newblock \doi{10.1007/s12021-008-9032-z}.

\bibitem[{Cappuccio et~al.(2015)Cappuccio, Tieri and Castiglione}]{CTC15}
Cappuccio A, Tieri P and Castiglione F (2015) {Multiscale modelling in
  immunology: a review}.
\newblock \emph{Briefings in Bioinformatics} 17(3): 408--418.
\newblock \doi{10.1093/bib/bbv012}.

\bibitem[{Center(2024)}]{apinatomyMap}
Center SH (2024) {Anatomical Flatmap Resources}.
\newblock
  \urlprefix\url{https://docs.sparc.science/docs/anatomical-flatmap-resources}.
\newblock Accessed: 2024-10-01.

\bibitem[{Dada and Mendes(2011)}]{DM+11}
Dada JO and Mendes P (2011) Multi-scale modelling and simulation in systems
  biology.
\newblock \emph{Integrative Biology} 3: 86--96.
\newblock \doi{10.1039/C0IB00075B}.

\bibitem[{de~Bono et~al.(2016)de~Bono, Helvensteijn, Kokash, Martorelli,
  Sarwar, Islam, Grenon and Hunter}]{roadmap2016}
de~Bono B, Helvensteijn M, Kokash N, Martorelli I, Sarwar D, Islam S, Grenon P
  and Hunter P (2016) Requirements for the formal representation of
  pathophysiology mechanisms by clinicians.
\newblock \emph{Interface Focus} 6.
\newblock \doi{10.1098/rsfs.2015.0099}.

\bibitem[{de~Bono et~al.(2017)de~Bono, Safaei, Grenon and Hunter}]{BSG+17}
de~Bono B, Safaei S, Grenon P and Hunter PJ (2017) {Meeting the multiscale
  challenge: representing physiology processes over ApiNATOMY circuits using
  bond graphs}.
\newblock \emph{Interface focus} 8.
\newblock \doi{10.1098/rsfs.2017.0026}.

\bibitem[{de~Bono et~al.(2015)de~Bono, Safaei, Grenon, Nickerson, Alexander,
  Helvensteijn, Kok, Kokash, Wu, Yu, Hunter and Baldock}]{BSG+15}
de~Bono B, Safaei S, Grenon P, Nickerson DP, Alexander S, Helvensteijn M, Kok
  JN, Kokash N, Wu A, Yu T, Hunter P and Baldock RA (2015) The open physiology
  workflow: modeling processes over physiology circuitboards of interoperable
  tissue units.
\newblock \emph{Frontiers in Physiology} 6: 24.
\newblock \doi{10.3389/fphys.2015.00024}.

\bibitem[{de~Matos et~al.(2010)de~Matos, Dekker, Ennis, Hastings, Haug, Turner
  and Steinbeck}]{CHEBI}
de~Matos P, Dekker A, Ennis M, Hastings J, Haug K, Turner S and Steinbeck C
  (2010) Chebi: a chemistry ontology and database.
\newblock \emph{Journal of Cheminformatics} 2(1).
\newblock \doi{10.1186/1758-2946-2-S1-P6}.

\bibitem[{Deisboeck et~al.(2011)Deisboeck, Wang, Macklin and Cristini}]{DWZ+11}
Deisboeck TS, Wang Z, Macklin P and Cristini V (2011) Multiscale cancer
  modeling.
\newblock \emph{Annual Review of Biomedical Engineering} 13: 127--55.
\newblock \doi{10.1146/annurev-bioeng-071910-124729}.

\bibitem[{Diehl et~al.(2016)Diehl, Meehan, Bradford and et~al.}]{CL}
Diehl A, Meehan T, Bradford YM and et~al (2016) {The Cell Ontology 2016:
  enhanced content, modularization, and ontology interoperability}.
\newblock \emph{Journal of biomedical semantics} 7.
\newblock \doi{10.1186/s13326-016-0088-7}.

\bibitem[{Dud{\'a}{\u s} et~al.(2018)Dud{\'a}{\u s}, Lohmann, Svátek and
  Pavlov}]{ontoVis-survey}
Dud{\'a}{\u s} M, Lohmann S, Svátek V and Pavlov D (2018) Ontology
  visualization methods and tools: a survey of the state of the art.
\newblock \emph{The Knowledge Engineering Review} 33.
\newblock \doi{10.1017/S0269888918000073}.

\bibitem[{Eissing et~al.(2011)Eissing, Kuepfer, Becker, Block and
  et~al.}]{EKB+11}
Eissing T, Kuepfer L, Becker C, Block MS and et~al (2011) A computational
  systems biology software platform for multiscale modeling and simulation:
  Integrating whole-body physiology, disease biology, and molecular reaction
  networks.
\newblock \emph{Frontiers in Physiology} \doi{10.3389/fphys.2011.00004}.

\bibitem[{Filice and Kahn(2021)}]{filice2021biomedical}
Filice RW and Kahn CE (2021) Biomedical ontologies to guide ai development in
  radiology.
\newblock \emph{Journal of Digital Imaging} 34(6): 1331--1341.
\newblock \doi{10.1007/s10278-021-00527-1}.

\bibitem[{Gardner et~al.(2008)Gardner, Akil, Ascoli, Bowden, Bug, Donohue,
  Goldberg, Grafstein, Grethe, Gupta et~al.}]{gardner2008neuroscience}
Gardner D, Akil H, Ascoli G, Bowden D, Bug W, Donohue D, Goldberg D, Grafstein
  B, Grethe J, Gupta A et~al. (2008) The neuroscience information framework: a
  data and knowledge environment for neuroscience.
\newblock \emph{Neuroinformatics} 6(3): 149--160.
\newblock \doi{10.1007/s12021-008-9024-z}.

\bibitem[{Gaudet et~al.(2017)Gaudet, {\v{S}}kunca, Hu and Dessimoz}]{GO}
Gaudet P, {\v{S}}kunca N, Hu JC and Dessimoz C (2017) Primer on the gene
  ontology.
\newblock In: \emph{The Gene Ontology Handbook}. Springer, pp. 25--37.
\newblock \urlprefix\url{https://doi.org/10.1007/978-1-4939-3743-1_3}.

\bibitem[{Gillespie et~al.(2022)Gillespie, Tripathy, Sy
  et~al.}]{gillespie2022neuron}
Gillespie TH, Tripathy SJ, Sy MF et~al. (2022) The neuron phenotype ontology: A
  fair approach to proposing and classifying neuronal types.
\newblock \emph{Neuroinformatics} 20: 793--809.
\newblock \doi{10.1007/s12021-022-09566-7}.

\bibitem[{Grethe et~al.(2014)Grethe, Bandrowski, Davis, Christopher, Gupta,
  Larson, Yueling, Ibrahim, Andrea, Patricia, Marenco, Perry, Rixin, Gordon and
  Martone}]{SciCrunch}
Grethe J, Bandrowski A, Davis B, Christopher C, Gupta A, Larson S, Yueling L,
  Ibrahim O, Andrea S, Patricia W, Marenco L, Perry M, Rixin W, Gordon S and
  Martone M (2014) {SciCrunch: A cooperative and collaborative data and
  resource discovery platform for scientific communities}.
\newblock \emph{Frontiers in Neuroinformatics} 8.
\newblock \doi{10.3389/conf.fninf.2014.18.00069}.

\bibitem[{Group(2024)}]{ApiNATOMY-models}
Group SAW (2024) {Catalogue of ApiNATOMY connectivity models}.
\newblock \urlprefix\url{https://scicrunch.org/sawg/about/ApiNATOMY}.
\newblock Accessed: 2024-10-01.

\bibitem[{Hayamizu et~al.(2013)Hayamizu, Mangan, Corradi, Kadin and
  Ringwald}]{hayamizu2013mouse}
Hayamizu TF, Mangan ME, Corradi JP, Kadin JA and Ringwald M (2013) {The Adult
  Mouse Anatomical Dictionary: A tool for annotating and integrating data}.
\newblock \emph{Mammalian Genome} 24(9): 422--430.
\newblock \doi{10.1186/gb-2005-6-3-r29}.

\bibitem[{He et~al.(2023)He, Liu, Yang, Hannink, Hammer, Popescu and
  Xu}]{he2023applications}
He F, Liu K, Yang Z, Hannink M, Hammer RD, Popescu M and Xu D (2023)
  Applications of cutting-edge artificial intelligence technologies in
  biomedical literature and document mining.
\newblock \emph{Medical Review} 3(3): 200--204.
\newblock \doi{10.1515/mr-2023-0011}.

\bibitem[{Hoehndorf et~al.(2015)Hoehndorf, Schofield and
  Gkoutos}]{hoehndorf2015role}
Hoehndorf R, Schofield PN and Gkoutos GV (2015) The role of ontologies in
  biological and biomedical research: a functional perspective.
\newblock \emph{Briefings in Bioinformatics} 16(6): 1069--1080.
\newblock \doi{10.1093/bib/bbv011}.

\bibitem[{Hunter et~al.(2002)Hunter, Robbins and Noble}]{HRN02}
Hunter P, Robbins P and Noble D (2002) The iups human physiome project.
\newblock \emph{Pfl{\"u}gers Archiv} 445(1): 1--9.
\newblock \urlprefix\url{https://doi.org/10.1007/s00424-002-0890-1}.

\bibitem[{IETF Working Group()}]{JSON}
IETF Working Group (2024) {JSON Schema}.
\newblock \urlprefix\url{https://json-schema.org}.
\newblock Accessed: 2024-10-01.

\bibitem[{Kobourov(2013)}]{kobourov2013force}
Kobourov SG (2013) Force-directed drawing algorithms.
\newblock In: Tamassia R (ed.) \emph{Handbook of Graph Drawing and
  Visualization}, chapter~12. CRC Press, pp. 383--408.
\newblock \doi{10.1201/b15385}.

\bibitem[{Kokash(2024{\natexlab{a}})}]{ApiNATOMY-demo}
Kokash N (2024{\natexlab{a}}) Open physiology viewer.
\newblock \urlprefix\url{http://open-physiology-viewer.surge.sh}.
\newblock Accessed: 2024-10-01.

\bibitem[{Kokash(2024{\natexlab{b}})}]{ApiNATOMY-tool}
Kokash N (2024{\natexlab{b}}) Open physiology viewer source code.
\newblock
  \urlprefix\url{https://github.com/open-physiology/open-physiology-viewer}.
\newblock Accessed: 2024-10-01.

\bibitem[{Kokash and de~Bono(2021)}]{kokash2021knowledge}
Kokash N and de~Bono B (2021) {Knowledge Representation for Multi-Scale
  Physiology Route Modeling}.
\newblock \emph{Frontiers in Neuroinformatics} 15: 560050.
\newblock \doi{10.3389/fninf.2021.560050}.

\bibitem[{Lambrix et~al.(2007)Lambrix, Tan, Jakoniene and
  Str{\"o}mb{\"a}ck}]{bioOnto}
Lambrix P, Tan H, Jakoniene V and Str{\"o}mb{\"a}ck L (2007) {Biological
  Ontologies}.
\newblock In: \emph{Semantic Web: Revolutionizing Knowledge Discovery in the
  Life Sciences}. Springer, pp. 85--99.
\newblock \doi{10.1007/978-0-387-48438-9_5}.

\bibitem[{Mizeranschi et~al.(2015)Mizeranschi, Groen, Borgdorff, Hoekstra,
  Chopard and Dubitzky}]{MGB+15}
Mizeranschi A, Groen D, Borgdorff J, Hoekstra A, Chopard B and Dubitzky W
  (2015) Anatomy and physiology of multiscale modeling and simulation in
  systems medicine.
\newblock \emph{Methods in molecular biology (Clifton, N.J.)} 1386: 375--404.
\newblock \doi{10.1007/978-1-4939-3283-2_17}.

\bibitem[{Mungall et~al.(2012)Mungall, Torniai, Gkoutos, Lewis and
  Haendel}]{mungall2012uberon}
Mungall CJ, Torniai C, Gkoutos GV, Lewis SE and Haendel MA (2012) Uberon, an
  integrative multi-species anatomy ontology.
\newblock \emph{Genome Biology} 13(1): R5.
\newblock \doi{10.1186/gb-2012-13-1-r5}.

\bibitem[{Mungall et~al.(2017)}]{mungall2017monarch}
Mungall CJ et~al. (2017) The monarch initiative: an integrative data and
  analytic platform connecting phenotypes to genotypes across species.
\newblock \emph{Nucleic Acids Research} 45(D1): D712--D722.
\newblock \doi{10.1093/nar/gkw1128}.

\bibitem[{Musen(2015)}]{musen2015protege}
Musen MA (2015) The protégé project: A look back and a look forward.
\newblock \emph{AI Matters} 1(4): 4--12.
\newblock \doi{10.1145/2757001.2757003}.

\bibitem[{Nickerson et~al.(2023)Nickerson, Brooks, Balachandran, Gillespie,
  Imam, Grethe, Tappan, de~Bono, Martone and Hunter}]{flatmaps}
Nickerson D, Brooks D, Balachandran B, Gillespie T, Imam F, Grethe J, Tappan S,
  de~Bono B, Martone M and Hunter P (2023) {Generating interactive visual maps
  of anatomical connectivity from SPARC connectivity knowledge}.
\newblock \emph{Physiology} 38.
\newblock \doi{10.1152/physiol.2023.38.S1.5796152}.

\bibitem[{Noy et~al.(2009)Noy, Shah, Whetzel, Dai, Dorf, Griffith, Jonquet,
  Rubin, Storey, Chute and Musen}]{noy2009bioportal}
Noy NF, Shah NH, Whetzel PL, Dai B, Dorf M, Griffith N, Jonquet C, Rubin DL,
  Storey MA, Chute CG and Musen MA (2009) Bioportal: ontologies and integrated
  data resources at the click of a mouse.
\newblock \emph{Nucleic Acids Research} 37(Web Server issue): W170--W173.
\newblock \doi{10.1093/nar/gkp440}.

\bibitem[{Osanlouy et~al.(2021)Osanlouy, Bandrowski, de~Bono and
  et~al.}]{osanlouy2021sparc}
Osanlouy M, Bandrowski A, de~Bono B and et~al (2021) {The SPARC DRC: Building a
  Resource for the Autonomic Nervous System Community}.
\newblock \emph{Frontiers in Physiology} 12: 693735.
\newblock \doi{10.3389/fphys.2021.693735}.

\bibitem[{{Rassingthwaighte} and {Chizeck}(2008)}]{RC08}
{Rassingthwaighte} JB and {Chizeck} HJ (2008) The physiome projects and
  multiscale modeling.
\newblock \emph{IEEE Signal Processing Magazine} 25(2): 121--144.
\newblock \doi{10.1109/MSP.2007.914723}.

\bibitem[{Robinson et~al.(2008)}]{robinson2008human}
Robinson PN et~al. (2008) {The Human Phenotype Ontology: A Tool for Annotating
  and Analyzing Human Hereditary Disease}.
\newblock \emph{American Journal of Human Genetics} 83(5): 610--615.
\newblock \doi{10.1016/j.ajhg.2008.09.017}.

\bibitem[{Rosse and Mejino(2008)}]{FMA}
Rosse C and Mejino J (2008) {The Foundational Model of Anatomy Ontology}.
\newblock \emph{Anatomy Ontologies for Bioinformatics: Principles and Practice}
  6.
\newblock \doi{10.1007/978-1-84628-885-2_4}.

\bibitem[{Savova et~al.(2019)Savova, Danciu, Alamudun, Miller, Lin, Bitterman,
  Tourassi and Warner}]{savova2019use}
Savova GK, Danciu I, Alamudun F, Miller T, Lin CH, Bitterman DS, Tourassi G and
  Warner JL (2019) Use of natural language processing to extract clinical
  cancer phenotypes from electronic medical records.
\newblock \emph{Cancer Research} 79(21): 5463--5470.
\newblock \doi{10.1158/0008-5472.CAN-19-0579}.

\bibitem[{Schriml et~al.(2012)}]{schriml2012disease}
Schriml LM et~al. (2012) Disease ontology: a backbone for disease semantic
  integration.
\newblock \emph{Nucleic Acids Research} 40(D1): D940--D946.
\newblock \doi{10.1093/nar/gkr972}.

\bibitem[{SciGraph()}]{SciGraph}
SciGraph (2024) {SciGraph}.
\newblock \urlprefix\url{https://github.com/SciGraph/SciGraph}.
\newblock Accessed: 2024-10-01.

\bibitem[{Shannon et~al.(2003)}]{shannon2003cytoscape}
Shannon P et~al. (2003) Cytoscape: A software environment for integrated models
  of biomolecular interaction networks.
\newblock \emph{Genome Research} 13(11): 2498--2504.
\newblock \doi{10.1101/gr.1239303}.

\bibitem[{Smith et~al.(2007)}]{smith2007obo}
Smith B et~al. (2007) {The OBO Foundry: coordinated evolution of ontologies to
  support biomedical data integration}.
\newblock \emph{Nature Biotechnology} 25(11): 1251--1255.
\newblock \doi{10.1038/nbt1346}.

\bibitem[{Snyder et~al.(2019)Snyder, Lin, Posgai and et~al.}]{HuBMAP}
Snyder MP, Lin S, Posgai A and et~al (2019) Mapping the human body at cellular
  resolution -- the {NIH Common Fund Human BioMolecular Atlas Program}.
\newblock \emph{arXiv e-prints} \doi{10.1038/s41586-019-1629-x}.

\bibitem[{SPARC()}]{Sparc}
SPARC (2024) {Stimulating Peripheral Activity to Relieve Conditions}.
\newblock \urlprefix\url{https://commonfund.nih.gov/sparc}.
\newblock Accessed: 2024-10-01.

\bibitem[{Surles-Zeigler et~al.(2021)Surles-Zeigler, Sincomb, Gillespie,
  de~Bono, Bresnahan, Mawe, Grethe, Hendricks~Tappan, Heal and
  Martone}]{surles2022extending}
Surles-Zeigler M, Sincomb T, Gillespie T, de~Bono B, Bresnahan J, Mawe G,
  Grethe J, Hendricks~Tappan S, Heal M and Martone M (2021) {Extending and
  using anatomical vocabularies in the Stimulating Peripheral Activity to
  Relieve Conditions (SPARC) project}.
\newblock \emph{bioRxiv} \doi{10.1101/2021.11.15.467961}.

\bibitem[{Viceconti and Hunter(2016)}]{VH16}
Viceconti M and Hunter P (2016) The virtual physiological human: Ten years
  after.
\newblock \emph{Annual Review of Biomedical Engineering} 18: 103--123.
\newblock \doi{10.1146/annurev-bioeng-110915-114742}.

\bibitem[{W3C JSON-LD Working Group()}]{JSON-LD}
W3C JSON-LD Working Group (2024) {JSON for Linking Data}.
\newblock \urlprefix\url{https://json-ld.org/}.
\newblock Accessed: 2024-10-01.

\bibitem[{W3C Working Group()}]{RDF}
W3C Working Group (2024) {Resource Description Framework (RDF)}.
\newblock \urlprefix\url{https://www.w3.org/RDF/}.
\newblock Accessed: 2024-10-01.

\bibitem[{Walpole et~al.(2013)Walpole, Papin and Peirce}]{WPP13}
Walpole J, Papin JA and Peirce SM (2013) Multiscale computational models of
  complex biological systems.
\newblock \emph{Annual Review of Biomedical Engineering} 15.
\newblock \doi{10.1146/annurev-bioeng-071811-150104}.

\bibitem[{Wilkinson et~al.(2016)Wilkinson, Dumontier and
  Aalbersberg}]{wilkinson2016fair}
Wilkinson MD, Dumontier M and Aalbersberg IJea (2016) The fair guiding
  principles for scientific data management and stewardship.
\newblock \emph{Scientific Data} 3: 160018.
\newblock \doi{10.1038/sdata.2016.18}.

\bibitem[{Wolkenhauer et~al.(2014)Wolkenhauer, Auffray, Brass and
  et~al.}]{WAB+2014}
Wolkenhauer O, Auffray C, Brass O and et~al (2014) Enabling multiscale modeling
  in systems medicine.
\newblock \emph{Genome Medicine} 6(3).
\newblock \doi{10.1186/gm538}.

\end{thebibliography}

\end{document}